\documentclass[11pt]{article}

\usepackage[margin=1in]{geometry}
\usepackage{amsmath, amssymb}
\usepackage{booktabs}
\usepackage{graphicx}
\usepackage{xcolor}
\usepackage[colorlinks=true,linkcolor=blue!60!black,citecolor=blue!60!black,urlcolor=blue!60!black]{hyperref}
\usepackage{natbib}
\usepackage{caption}
\usepackage{array}
\usepackage[T1]{fontenc}
\usepackage{lmodern}
\usepackage{microtype}
\usepackage{enumitem}
\usepackage{setspace}
\usepackage{float}

\title{Spectral Probe-Circuits\\
\large A Three-Step Recipe for Identifying Attention-Head Circuits\\
in Pretrained Transformers\thanks{Correspondence: \texttt{abbyxu@gmail.com}. Code, data, and reproducibility scripts: \url{https://github.com/skydancerosel/spectral-probe-circuits}}}
\author{Yongzhong Xu}
\date{}

\newcommand{\PR}{\mathrm{PR}}

\begin{document}

\maketitle

\begin{abstract}
We present a three-step recipe for identifying attention-head circuits in pretrained transformers. (1)~A \emph{spectral signal} -- the time-integrated participation ratio (PR) of per-head attention output -- ranks heads doing sustained content-dependent computation without labels or attribution gradients. (2)~A \emph{task-pattern screen} applied over all heads filters that general indicator into a task-specific candidate circuit by measuring attention from the task-relevant query position to canonical target positions. (3)~A \emph{causal verification} step group-ablates the candidate circuit against a matched-random control in the same layers, plus an all-heads-in-layer upper bound. The PR-integral is a general specialization indicator; the screen is what makes it task-specific; the matched-random differential is what makes the causal claim honest.

We validate the recipe across an 8$\times$ parameter range (51M to 1B-active~/~7B-total) and across two architecture families (dense transformer, mixture-of-experts) and three pretraining pipelines (TinyStories with a key-retrieval probe, FineWeb, the Pile, DCLM). Three findings frame the paper. First, the recipe ports: a small (3--6 head) induction circuit is causally necessary in every model tested, identified by the same screen-and-ablate procedure. Second, the per-head PR signal is predictive: on six independent seeds of a 51M-parameter probe model, the same spectral computation identifies the seed-specific circuit on each seed without any task labels. Third, the fraction of heads doing identifiable specialized computation is conserved at $\sim$17--19\% across an 8$\times$ scale range, while the specific capability circuits themselves stay 3--11 heads -- sublinear in total head count.

This paper is the methodology anchor of a three-paper program. The companion papers extend the recipe to developmental trajectories during pretraining \citep{paper2_developmental} and to composed-task circuits where pattern selectivity decouples from task-causal structure \citep{paper3_circuits}. Both companions cite the recipe established here.
\end{abstract}

\section{Introduction}

Mechanistic interpretability typically identifies attention-head circuits after they emerge, by ablating heads on a fully-trained model with a fixed target capability and inspecting the ones whose removal degrades it. That workflow has produced detailed circuit-level accounts of small models \citep{olsson2022induction, wang2022ioi, conmy2023acdc, hanna2023greater, gould2023successor}, but it is expensive -- every condition is a forward pass over a fully-trained network -- and post hoc, in the sense that the target capability must be specified before the circuit search begins.

This paper develops a complementary methodology: a \emph{per-head spectral signal} that can be read off during training, plus an \emph{all-head task-pattern screen} that specializes that signal to a target task, plus standard causal ablation. The signal does not require labels or attribution gradients; it is the participation ratio of a per-head activation matrix, integrated over training. The screen is a fixed six-class set of canonical attention patterns (induction, previous-token, duplicate-token, first-token/BOS, self, local) plus task-specific patterns when they are needed. Causal verification follows the standard ablation paradigm with two controls: a matched-random ablation in the same layers, and an all-heads-in-layer upper bound.

We make three contributions:

\begin{itemize}[leftmargin=*,topsep=2pt,itemsep=2pt]
\item \textbf{A specific recipe with explicit definitions and thresholds} (Section~\ref{sec:method}), validated by replication across seven model configurations spanning 51M to 7B parameters.
\item \textbf{An honest framing of pattern-vs-task-causal status}. The PR-integral is a general specialization indicator: in attention-sink-dominated 1B-class models, top-$K$ by PR-integral surfaces L0/L1 generic content-dependent heads ahead of the task-causal circuit. The task-pattern screen is what makes the procedure task-specific; the matched-random differential is what makes the causal claim falsifiable. We state this in the methodology section (\S\ref{sec:framing}), not buried as a limitation.
\item \textbf{Cross-architecture replication and a conserved fraction}. A 3--6 head induction circuit is causally necessary in every model tested. The fraction of heads doing identifiable specialized work stays in the 17--19\% band across an 8$\times$ scale range, while the specific capability circuits themselves stay 3--11 heads -- sublinear in total head count.
\end{itemize}

\paragraph{Trilogy scope.} This is Paper~1 of a three-paper program. Paper~2 extends the recipe to developmental trajectories (the spectral signal precedes capability-selectivity emergence at intermediate training checkpoints) \citep{paper2_developmental}. Paper~3 applies the recipe to composed tasks (Indirect Object Identification, greater-than, successor sequences, variable binding) and documents a pattern-vs-task-causal decoupling whereby the same task is implemented by different primary attention patterns across model families \citep{paper3_circuits}. The current paper establishes the recipe and the cross-architecture induction results that both companions build on.

\begin{figure}[H]
  \centering
  \includegraphics[width=0.95\linewidth]{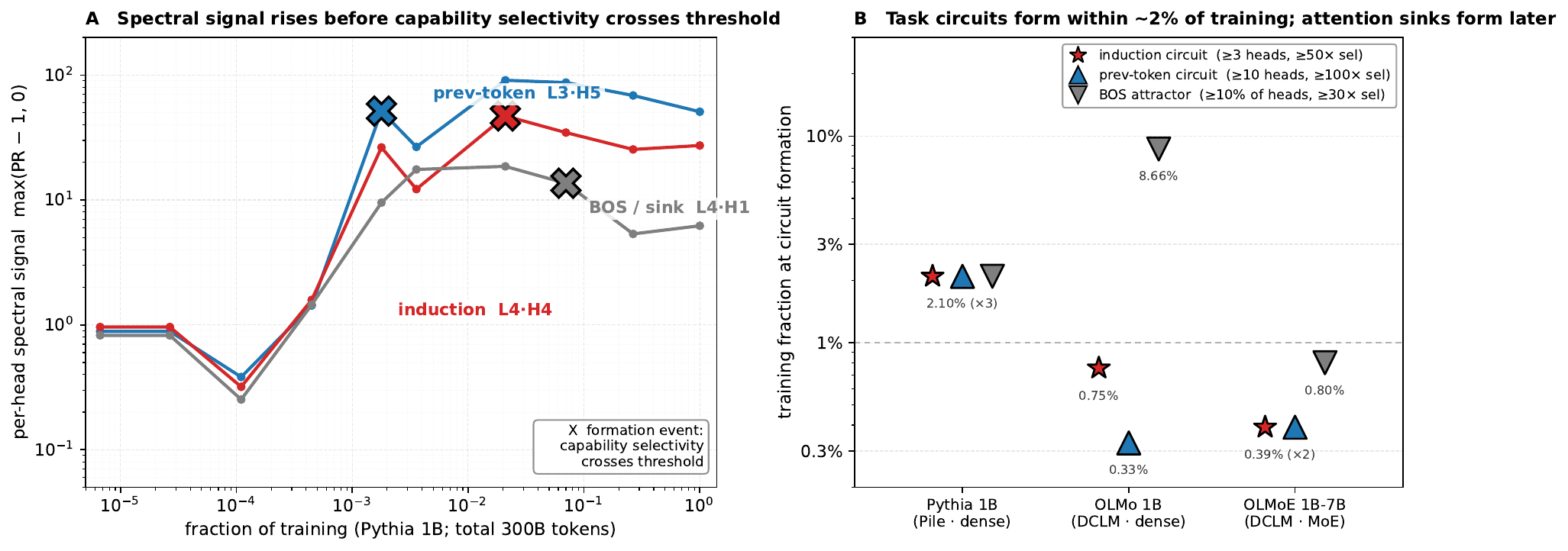}
  \caption{\textbf{Capability circuits emerge early in pretraining and the per-head spectral signal precedes their formation.} \textbf{(A)}~Per-head spectral signal at each checkpoint -- $\max(\PR_t - 1, 0)$, the integrand of the PR-integral ranking statistic defined in \S\ref{sec:step1}, plotted per-checkpoint rather than as a cumulative sum so the temporal structure of emergence is visible. Three identified heads in Pythia~1B: the induction head L4$\cdot$H4, the previous-token head L3$\cdot$H5, and a BOS / attention-sink head L4$\cdot$H1. X markers indicate the formation event, defined as the first training checkpoint at which the head's capability-selectivity ratio exceeds its threshold ($\geq 50\times$ for induction, $\geq 100\times$ for previous-token, $\geq 30\times$ for first-token). The spectral signal is elevated at or before formation in all three heads. \textbf{(B)}~Training fraction at circuit formation across three 1B-class configurations: Pythia~1B (Pile, dense), OLMo~1B-0724-hf (DCLM, dense), OLMoE 1B-7B-0924 (DCLM, MoE). Task circuits (induction, previous-token) form within the first 0.3--2.1\% of training in every configuration; the BOS attractor forms later. Companion paper~\citep{paper2_developmental} contains the full developmental analysis.}
  \label{fig:headline}
\end{figure}

\section{Related Work}

\paragraph{Induction heads and in-context learning.} \citet{olsson2022induction} identified induction heads -- attention heads implementing the \texttt{A B \dots\ A $\to$ B} copy pattern -- and connected their emergence to the in-context-learning phase transition during pretraining. We use the same operational definition (attention from the second-A position to the position after the first-A) but identify circuits differently: by a per-head spectral signal during training plus an all-head selectivity screen at any chosen checkpoint, replacing the integrated-gradients-style attribution they used.

\paragraph{Indirect Object Identification (IOI).} \citet{wang2022ioi} characterized the IOI circuit in GPT-2-small and identified head classes (name-movers, S-Inhibition, prev-token, induction, duplicate-token, negative name-movers, backup name-movers) and their compositional structure. Their decomposition is the canonical reference for composed-task circuits. Paper~3 of this program tests how much of that decomposition ports to three 1B-class models from different training pipelines \citep{paper3_circuits}; the present paper focuses on the recipe itself and on single-pattern capabilities (induction, previous-token) where it is cleanest.

\paragraph{Attention sinks.} \citet{xiao2024streaming} introduced the term ``attention sink'' for the empirical observation that pretrained LMs allocate large attention probability to the first token regardless of content, enabling streaming inference via KV-cache compression. We document that BOS-class heads (heads whose dominant attention target is the first position at $\geq 30\times$ over baseline) are widespread across 100M+-scale decoder-only LMs but absent from L0 and L1 at every training checkpoint sampled across all models we test. The companion paper \citep{paper2_developmental} characterizes BOS-fraction scaling with training data and architecture.

\paragraph{Automated circuit discovery.} \citet{conmy2023acdc} developed ACDC (Automatic Circuit DisCovery), an iterative edge-pruning algorithm for circuit identification given a fixed task and model. The recipe here is complementary: ACDC requires a fully-trained model and a defined task; the spectral-and-screen recipe operates per head, uses training-time signal (Step~1), and is intended as a fast pre-filter before more expensive analyses. The two methods address different stages of the interpretability workflow.

\paragraph{Participation ratio.} Participation ratio (effective rank) is a standard tool in random matrix theory and condensed matter physics for measuring the effective dimensionality of a distribution. Its use in interpretability is less established. The trajectory feature $I(L,H) = \sum_t \max(\PR_t - 1, 0)\,\Delta\log(\text{tokens}_t)$ defined in \S\ref{sec:step1} is the operative innovation rather than the bare PR; \S\ref{sec:integral_vs_spread} shows it beats every alternative trajectory feature we tested.

\paragraph{Cross-architecture mechanistic transfer.} Prior work documents that different model families produce different specific circuits for the same task \citep{lieberum2023chinchilla, marks2024sparsefeature}. We replicate that observation for induction at 1B-class scale (Section~\ref{sec:cross_arch}) and the companion paper sharpens it for composed tasks \citep{paper3_circuits}.

\paragraph{Spectral training dynamics.} Related work characterizes optimizer-induced drift in low dimensions \citep{xu2026optimizer} and signal--noise spectral edges in training trajectories \citep{xu2026spectraledge}. Those analyses look at training-trajectory geometry at the parameter level; ours looks at per-head activation spectra at fixed checkpoints. The connection is that high-dimensional content-dependent representations require many active singular directions, which is what the participation ratio measures.

\section{Methodology}
\label{sec:method}

\subsection{Panel of models}
\label{sec:setup}

Seven model configurations (six unique base models plus six TS-51M seeds):

\begin{center}
\footnotesize
\setlength{\tabcolsep}{4pt}
\begin{tabular}{lllll}
\toprule
Model & Scale & Architecture & Training data & Source \\
\midrule
TS-51M      & 51M     & GPT-2 8L/512d/16h & TinyStories + probe & this work, 6 seeds \\
GPT-2 124M  & 124M    & GPT-2 12L/768d/12h & FineWeb-10B & nanoGPT \citep{karpathy2023nanogpt} \\
Pythia 160M & 160M    & GPT-NeoX dense & Pile & \citet{biderman2023pythia} \\
Pythia 410M & 410M    & GPT-NeoX dense & Pile & \citet{biderman2023pythia} \\
Pythia 1B   & 1B      & GPT-NeoX dense & Pile & \citet{biderman2023pythia} \\
OLMo 1B     & 1B      & Llama-style dense & DCLM & \citet{groeneveld2024olmo} \\
OLMoE 1B-7B & 1B/7B   & Llama-style MoE (top-8/64) & DCLM & \citet{muennighoff2024olmoe} \\
\bottomrule
\end{tabular}
\end{center}

\paragraph{Synthetic induction batch.} 2000 sequences of length 256, RNG seed 42. Each sequence has structure \texttt{[filler] A B [filler] A}, where A and B are random tokens drawn from vocabulary IDs $[100, 10000)$. The induction prediction is B at the position immediately following the second occurrence of A.

\paragraph{TS-51M key-retrieval probe.} Sequences of the form
\begin{quote}\small
\texttt{[prefix] The secret code is XXXX. [filler] What is the secret code? $\to$ XXXX}
\end{quote}
where \texttt{XXXX} is a single-token codeword from a fixed 512-codeword vocabulary. Used in place of the synthetic induction batch for the 51M-scale model, which is too small for induction-batch evaluation to discriminate.

\paragraph{Natural-text batches.} OpenWebText sequences filtered to positions where the next ground-truth token has appeared earlier in the context (induction-target positions); used for the natural-text confirmation in \S\ref{sec:karpathy} and beyond.

\paragraph{Inference setup.} fp16 by default; fp32 used for Pythia~1B's early-checkpoint mech-interp (Section~\ref{sec:cross_arch}) because baseline attention values underflow fp16 representable range at random-init scale. Per-(layer, head) hooks placed at the attention output projection (\texttt{attention.dense} for GPT-NeoX models; \texttt{self\_attn.o\_proj} for Llama/OLMo/OLMoE).

\subsection{Step 1: Per-head spectral signal}
\label{sec:step1}

For each (layer $L$, head $H$) and training checkpoint $t$:

\begin{enumerate}[leftmargin=*,topsep=2pt,itemsep=2pt]
\item Extract the per-head attention output at the task-relevant query position over the fixed evaluation batch. The result is an activation matrix $M \in \mathbb{R}^{N \times d_{\mathrm{head}}}$ where $N$ is the batch size and $d_{\mathrm{head}}$ is per-head dimension.
\item Compute the singular value spectrum $\{\sigma_i\}$ of $M$.
\item Compute the participation ratio
\begin{equation}
\PR(L,H,t) \;=\; \exp\!\big(\mathcal{H}(p)\big), \qquad p_i = \sigma_i^2 \Big/ \sum_j \sigma_j^2,
\label{eq:pr}
\end{equation}
where $\mathcal{H}(p) = -\sum_i p_i \log p_i$ is the entropy of the squared-singular-value distribution.
\end{enumerate}

The trajectory feature is the time integral
\begin{equation}
I(L,H) \;=\; \sum_{t} \max\!\big(\PR(L,H,t) - 1,\ 0\big)\cdot \Delta \log(\text{tokens}_t).
\label{eq:integral}
\end{equation}

$I(L,H)$ weights \emph{sustained} content-dependent computation. Intuitively, a head whose per-position attention output is concentrated in one direction across the batch has rank $\approx 1$ (PR $\approx 1$, content-independent); a head whose output spans many directions (one direction per content variation in the batch) has high PR. The $\max(\PR - 1, 0)$ clipping prevents noisy random-init PR from dominating the integral.

\paragraph{Why integral, not PR-spread.} On Pythia models, L0 heads start at PR $\approx 60$ (random attention at initialization spreads probability across all positions, producing a high effective rank) and \emph{collapse} to PR $\approx 2$--$30$ by the end of training. Ranking heads by PR-spread (max minus min over training) flags these collapsing heads as top picks; ranking by $I(L,H)$ correctly demotes them in favor of heads that \emph{gain} sustained PR. The full nine-feature comparison appears in \S\ref{sec:integral_vs_spread}.

\subsection{Step 2: Task-pattern screen}
\label{sec:step2}

For each head, measure attention from the task-relevant query position to canonical target positions. Six standard classes are computed in all evaluations:

\begin{itemize}[leftmargin=*,topsep=2pt,itemsep=2pt]
\item \textbf{Induction} -- attention to the position immediately following the earlier occurrence of the current token (the B position in \texttt{A B \dots\ A $\to$ B}).
\item \textbf{Previous-token} -- attention to position $t-1$.
\item \textbf{Duplicate-token} -- attention to an earlier occurrence of the current token.
\item \textbf{First-token / BOS} -- attention to position 0.
\item \textbf{Self} -- attention to the current position.
\item \textbf{Local} -- mean attention over positions $t-2$ through $t-5$.
\end{itemize}

\paragraph{Selectivity.} Selectivity is the ratio of attention to the target positions over a uniform-other baseline:
\begin{equation}
\mathrm{sel}_{\mathrm{class}}(L,H) \;=\; \frac{\text{attention to target positions}}{1/(T - k)},
\end{equation}
where $T$ is the number of tokens in context and $k$ is the number of target positions for the class. Two thresholds:

\begin{itemize}[leftmargin=*,topsep=2pt,itemsep=2pt]
\item $\geq 30\times$ for class assignment. A head is classified into the class with maximum selectivity, provided that maximum exceeds 30$\times$.
\item $\geq 50\times$ for circuit membership (induction ablation set selection). $\geq 100\times$ for previous-token circuit (because absolute prev-token selectivity values are much larger -- some heads reach $10^6$--$10^7$).
\end{itemize}

\paragraph{All-head versus best-class.} Two analysis modes are used:

\begin{itemize}[leftmargin=*,topsep=2pt,itemsep=2pt]
\item \emph{Best-class ranking} -- for each head, take the highest-selectivity class as the head's ``type.'' Useful when capability classes are roughly balanced.
\item \emph{All-head capability-specific screen} -- for a single target capability $X$, take all heads with $\mathrm{sel}_X \geq$ threshold regardless of best class. Required in the attention-sink-dominated regime ($\geq 70\%$ of heads classify as first-token at the standard threshold; Pythia 410M and 1B-class models), where best-class ranking surfaces BOS-class heads ahead of capability-specific heads even when the capability heads have meaningful selectivity.
\end{itemize}

\subsection{Step 3: Causal verification}
\label{sec:step3}

Group-ablate the screen-identified circuit by zeroing the per-head slice of the residual contribution at the attention output projection. The hook is a forward pre-hook on the projection module; it zeros columns $[h\cdot d_{\mathrm{head}}\,{:}\,(h{+}1)\cdot d_{\mathrm{head}}]$ for each head $h$ in the ablation set.

Two controls per condition:

\begin{itemize}[leftmargin=*,topsep=2pt,itemsep=2pt]
\item \textbf{Matched-random.} Same layers as the circuit picks, equal head count per layer, no overlap with the picks. Controls for the layer composition of the ablation -- different layers contribute different amounts to the final logits.
\item \textbf{Upper bound.} All heads in the pick layers. Saturates the layer-level effect, so a ``true'' causal head must produce a similarly-large effect with many fewer heads ablated.
\end{itemize}

\paragraph{Metrics.} Top-1 accuracy and top-5 accuracy on the synthetic eval batch, plus per-example mean logit of the target token. When the matched-random effect is large or counter-intuitive, multi-seed sampling ($\geq 5$ seeds) is recommended; in the cases reported here it is small and one-sided, where single-seed estimates are adequate.

\subsection{Threshold calibration}
\label{sec:threshold_calibration}

The two free hyperparameters of the recipe -- $K$ (top-$K$ cutoff for the PR-integral spectral signal) and $T$ (selectivity threshold for circuit membership) -- were initially calibrated on a single model and transplanted as defaults across the cross-architecture panel. $K$ follows the linear-scaling rule $K \approx 0.18 \times n_{\mathrm{heads}}$ (the conserved-fraction observation of \S\ref{sec:conserved_fraction}, validated on Pythia~124M\,/\,160M\,/\,410M). $T = 50\times$ came from the Pythia~410M ablation-floor sweep (\S\ref{sec:distribution_vs_dilution}): ablating all heads with induction-selectivity $\geq 50\times$ drove induction performance to 0\%. Both were transplanted to the 1B panel without per-model re-validation.

\paragraph{Per-model ablation-floor re-validation.} A direct sweep on each of Pythia~1B, OLMo~1B, and OLMoE~1B-7B at $T \in \{2, 10, 30, 50, 100\}\times$ confirms the 50$\times$ transplant:

\begin{center}
\footnotesize
\begin{tabular}{lrlll}
\toprule
Model & Baseline & Per-model $T^*$ & 50$\times$ catches & Verdict \\
\midrule
Pythia 1B   & 4.05\% & 10--30$\times$ (11--6 heads) & 94\% of effect & Slightly tight \\
OLMo 1B     & 1.00\% & $\geq 100\times$ (2 heads) & 95\% of effect & Conservative \\
OLMoE 1B-7B & 4.80\% & 30--50$\times$ (4 heads, plateau) & 100\% of effect & Exactly right \\
\bottomrule
\end{tabular}
\end{center}

The 50$\times$ default captures the full causal effect in OLMo~1B and OLMoE and 94\% of the effect in Pythia~1B; the missing 6\% in Pythia~1B comes from heads in the 10$\times$--50$\times$ band that are likely multi-role first-token + induction. The threshold is defensible as a uniform default across the panel, with the caveat that per-model $T^*$ differs and Pythia~1B's full causal circuit is 6 heads at $\geq 30\times$ rather than 3 at $\geq 50\times$.

\paragraph{Null-selectivity as a per-model noise floor.} For each model, induction-selectivity is also computed against a random non-special target position (the ``null''), drawn 500 times per model. The null distribution gives a per-model noise floor for selectivity. Within the Pythia natural-text family, the count of heads with induction-selectivity above null$_{p99}$ lands at 18.1\% (Pythia~160M) and 18.5\% (Pythia~410M) -- independent recovery of the 17--19\% conserved-fraction band of \S\ref{sec:conserved_fraction}, by a procedure that does not target that band. Across the 1B panel the fraction varies (Pythia~1B 25.8\%, OLMo~1B 4.7\%, OLMoE 27.3\%) and correlates with BOS-attractor dominance rather than with model scale; the conserved-fraction claim is best stated as within-family-and-scale rather than universal.

\paragraph{Pre-filter threshold.} A uniform $T_{\mathrm{filter}} = 2\times$ is defensible across the panel: above every model's null$_{p99}$, captures 100\% of heads with selectivity $\geq 10\times$ in all five panel models, and reduces downstream per-head analysis to roughly 21--38\% of total heads. This is the recall-prioritized threshold for downstream causal verification; $T = 50\times$ remains the precision-prioritized threshold for ablation-validated circuit-membership claims.

\subsection{Honest framing}
\label{sec:framing}

The recipe is a \emph{recipe}, not three independent claims. Each step does a specific job and is unsound on its own.

\begin{itemize}[leftmargin=*,topsep=2pt,itemsep=2pt]
\item \textbf{Step~1 (PR-integral)} is a \emph{general specialization indicator}. It surfaces heads doing sustained content-dependent computation, but in attention-sink-dominated 1B-class models the top-$K$ is dominated by L0/L1 generic content-dependent heads, not the task circuit. Used alone, it cannot distinguish induction from previous-token from duplicate-token specialization.

\item \textbf{Step~2 (task-pattern screen)} is what makes the procedure task-specific. The screen converts the general PR signal into a candidate circuit for a named capability by measuring attention from the task-relevant query position to canonical target positions. Used alone (without Step~1), it would face the multiple-testing problem of trying many candidate patterns; in practice we use Step~1 to constrain attention to specialization-rich heads first.

\item \textbf{Step~3 (causal verification)} is what makes the claim falsifiable. The matched-random control rules out the explanation ``ablating any 3--6 heads in those layers would have produced a similar effect''; the upper-bound control rules out the explanation ``the layers in question collectively matter and the screen-picked heads are just along for the ride.''
\end{itemize}

In what follows, when we say a circuit ``is causal,'' we mean: Step~3 shows the screen-picked heads produce an ablation effect substantially larger than matched-random in the same layers, while the upper bound saturates near-zero. When we say a circuit ``is identifiable from spectral signal alone,'' we mean: in Step~1's top-$K$, the screen-picked heads appear early in the integral ranking, and the recipe finds them without label supervision. When we say ``the recipe is predictive,'' we mean: applying Steps~1--2 to intermediate training checkpoints recovers most of the final-checkpoint circuit using a small fraction of training tokens, a claim developed in the companion paper~\citep{paper2_developmental}.

\section{TS-51M Six-Seed Validation}
\label{sec:ts51m}

The deepest causal validation of the recipe is on TS-51M, a 51M-parameter model pretrained six times with different random seeds (s42, s271, s149, s256, s123, s314) on TinyStories \citep{eldan2023tinystories} with periodic injection of a key-retrieval probe task. All six seeds learn the task; all six implement it with \emph{different} attention heads. The smallness of the model lets us run multi-seed ablations cheaply and characterize the cross-seed variance of the recipe, which would be cost-prohibitive at 1B+ scale.

\begin{figure}[H]
  \centering
  \includegraphics[width=\linewidth]{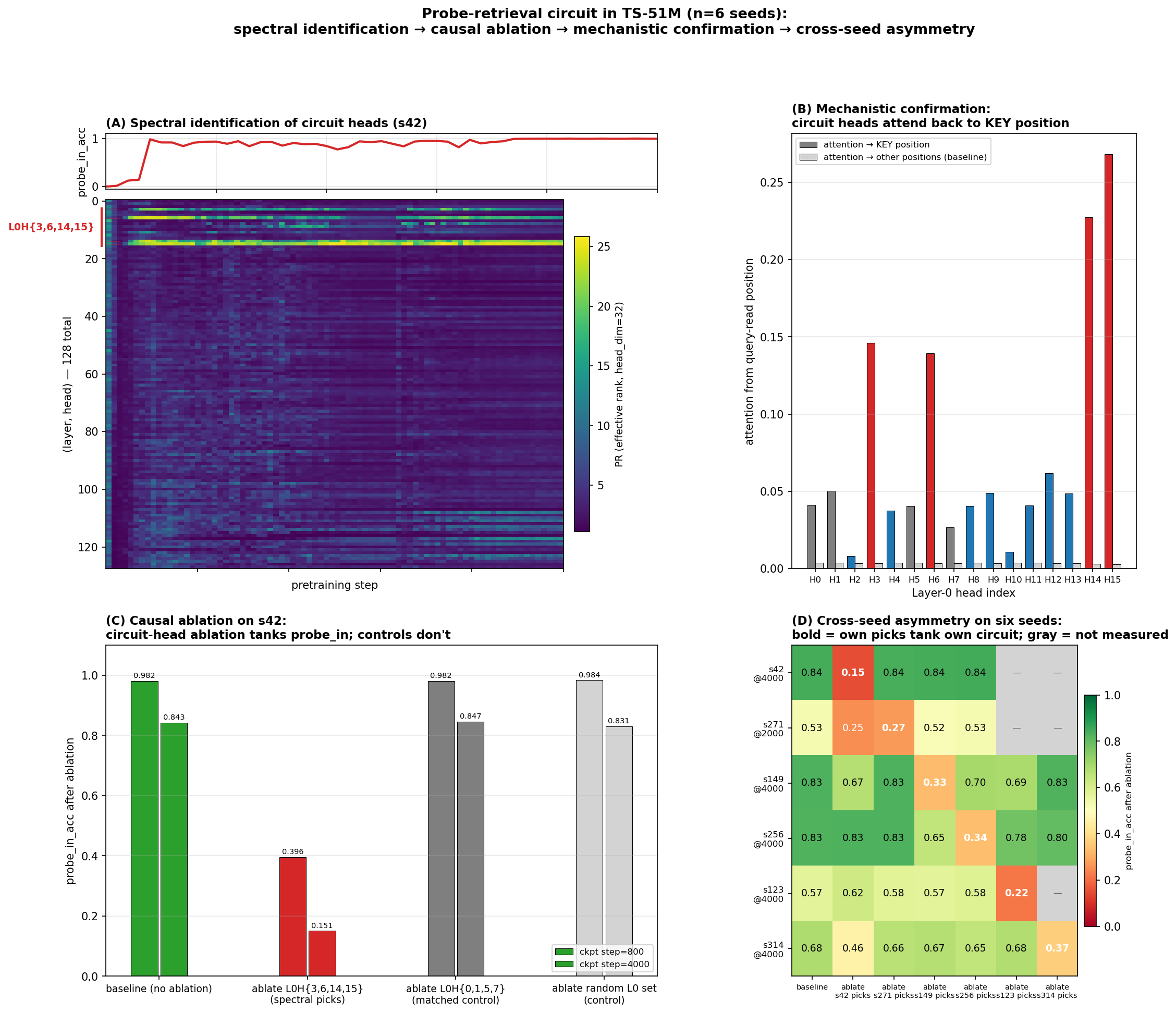}
  \caption{\textbf{The TS-51M six-seed validation, in four panels.} \textbf{(A)}~Per-head spectral signal (PR effective rank, $\max(\PR_t-1, 0)$) at each pretraining step for all 128 heads of TS-51M seed s42, plotted as a heatmap. The four L0 heads L0$\cdot$H\{3, 6, 14, 15\} (labeled in red) stand out from the rest of the model around step 800, coincident with the probe-task emergence event (top panel, \texttt{probe\_in\_acc}). \textbf{(B)}~Mechanistic confirmation: the four s42 circuit heads attend back to the KEY position (red bars) substantially more than they attend to other positions in the sequence (gray bars). The four heads identified by the spectral signal in (A) are doing the mechanism the task requires. \textbf{(C)}~Causal ablation on s42: ablating the four circuit heads (red bar) drops \texttt{probe\_in\_acc} from baseline $\sim 0.96$ at step 4000 to $\sim 0.20$; ablating a matched-random set of four heads in the same layer (L0) leaves accuracy at baseline; ablating ten random heads from L1--L7 produces no effect. Steps 800 (mid-emergence) and 4000 (fully trained) shown side by side. \textbf{(D)}~Cross-seed asymmetry: each seed's identified circuit (rows) is ablated under every seed's pretrained model (columns). The diagonal -- ablating a seed's own circuit on that seed's model -- tanks probe accuracy (dark red); off-diagonal cells where a different seed's heads are ablated on this seed's model show no effect (green). The six seeds use \emph{different} attention heads to solve the same task, and the spectral signal correctly identifies the seed-specific circuit on each seed without labels.}
  \label{fig:ts51m}
\end{figure}

\subsection{Setup}

\paragraph{Model.} TS-51M, 8 layers $\times$ 512-dim $\times$ 16 heads $\times$ 2048 FFN (b$_2$=0.95 for s42 pretraining).

\paragraph{Task.} TinyStories with key-retrieval probe injection at probe-rate 0.10. Each probe example has the structure \texttt{[prefix] [KEY codeword] [middle\dots] [QUERY codeword]}. At the QUERY position the model must retrieve the codeword introduced earlier in the prefix.

\paragraph{Eval set.} \texttt{probe\_eval\_in} -- 2000 probe-style examples with gap 5--30 between key and query. \texttt{probe\_eval\_ood} -- analogous but with longer gaps and distractor tokens.

\paragraph{Behavioral curve.} On s42, \texttt{probe\_in\_acc} rises from 0 to 0.05 at step~400, to 0.5 at step~800, and to 0.92 at step~1000.

\subsection{Per-head spectral on s42: four L0 standouts}
\label{sec:s42_per_head}

For each pretraining checkpoint $t$ and each (layer, head), we extract the attention output at the QUERY position of each probe example, forming a matrix of shape $[N_{\mathrm{examples}}{=}2000,\ d_{\mathrm{head}}{=}32]$. We compute its singular-value spectrum and the participation ratio (Eq.~\ref{eq:pr}).

Among the 128 (layer, head) pairs on s42, only four show a PR transition with spread $>19$ during the probe-emergence window (steps 200--1000):

\begin{center}
\small
\begin{tabular}{lrrrrr}
\toprule
Head & min PR & min @ step & max PR & max @ step & spread \\
\midrule
L0H14 & 1.63 & 400 & 24.40 & 800 & 22.8 \\
L0H6  & 2.51 & 400 & 24.72 & 800 & 22.2 \\
L0H15 & 2.74 & 400 & 24.11 & 1000 & 21.4 \\
L0H3  & 1.78 & 400 & 21.69 & 900 & 19.9 \\
L0H11 & 2.65 & 400 & 14.61 & 800 & 12.0 \\
(every other head, all layers) & --- & --- & --- & --- & $<10$ \\
\bottomrule
\end{tabular}
\end{center}

The four standout heads have PR $\approx 2$ at step~400 (when \texttt{probe\_in\_acc} just becomes nonzero) and PR $\approx 22$ by step~800. Late layers (L1--L7) do not show this pattern -- most heads have PR spreads below 10.

\textbf{Mechanistic intuition.} PR $\approx 2$ means the attention output across all 2000 probe examples is concentrated in essentially one direction -- the head is attending to a single ``default'' position regardless of the codeword in the prefix. PR $\approx 22$ means the output spans many directions, i.e.\ the head's attention has become content-dependent (attending to wherever the codeword was mentioned in each example), so its V-output varies with codeword identity.

\subsection{Causal ablation tables and the five concrete claims}
\label{sec:s42_ablation}

We zero out the per-head attention output (the input to the output projection for the ablated head columns) and re-evaluate \texttt{probe\_eval\_in}, \texttt{probe\_eval\_ood}, and plain LM \texttt{val\_loss}. We test at two checkpoints: step~800 (mid-grok, \texttt{probe\_in\_acc} $\approx 0.98$) and step~4000 (fully trained, \texttt{probe\_in\_acc} $\approx 0.84$).

Conditions: baseline; ablate the four candidate heads; ablate each candidate individually; ablate a matched-size control L0H\{0,1,5,7\}; ablate a random L0 set; ablate all 16 L0 heads.

\paragraph{Step~4000 (fully trained).}
\begin{center}
\small
\begin{tabular}{lrrr}
\toprule
Condition & probe\_in & probe\_ood & val\_loss \\
\midrule
baseline                              & \textbf{0.843} & 0.221 & 1.706 \\
\textbf{ablate L0H\{3,6,14,15\}}      & \textbf{0.151} $\downarrow$ & 0.129 & 1.823 \\
ablate L0H3 alone                     & 0.831 & 0.217 & 1.712 \\
ablate L0H6 alone                     & 0.807 & 0.210 & 1.740 \\
ablate L0H14 alone                    & 0.772 & 0.181 & 1.715 \\
ablate L0H15 alone                    & 0.596 & 0.179 & 1.725 \\
ablate L0H\{0,1,5,7\} (matched ctrl)  & 0.847 & 0.340 & 1.802 \\
ablate random L0 set                  & 0.831 & 0.204 & 1.754 \\
ablate all 16 L0 heads                & 0.129 & 0.132 & 3.963 \\
\bottomrule
\end{tabular}
\end{center}

\paragraph{Step~800 (mid-grok).}
\begin{center}
\small
\begin{tabular}{lrrr}
\toprule
Condition & probe\_in & probe\_ood & val\_loss \\
\midrule
baseline                              & \textbf{0.982} & 0.939 & 3.469 \\
\textbf{ablate L0H\{3,6,14,15\}}      & \textbf{0.396} $\downarrow$ & 0.187 $\downarrow$ & 3.567 \\
ablate L0H\{0,1,5,7\} (matched ctrl)  & 0.982 & 0.935 & 3.511 \\
ablate random L0 set                  & 0.984 & 0.943 & 3.509 \\
ablate all 16 L0 heads                & 0.135 & 0.139 & 4.101 \\
\bottomrule
\end{tabular}
\end{center}

\paragraph{Five concrete claims.}

\begin{enumerate}[leftmargin=*,topsep=2pt,itemsep=2pt]
\item \textbf{Specificity.} Ablating L0H\{3,6,14,15\} drops \texttt{probe\_in} by 0.69 at step~4000 (0.84 $\to$ 0.15) and by 0.59 at step~800 (0.98 $\to$ 0.40). Ablating any matched-size control set of L0 heads has near-zero effect.

\item \textbf{Redundancy.} No single head among the four is sufficient. The largest single-head ablation effect at step~4000 is L0H15 alone (\texttt{probe\_in} 0.84 $\to$ 0.60). The full effect requires removing all four.

\item \textbf{Other L0 heads carry general LM, not probe-task work.} Ablating all 16 L0 heads tanks \texttt{val\_loss} from 1.71 to 3.96 (catastrophic for general LM) but gives only a marginally larger \texttt{probe\_in} drop than ablating just the four (0.13 vs 0.15). The remaining 12 heads contribute essentially nothing to probe-task performance.

\item \textbf{Selectivity.} Ablating just the four circuit heads at step~800 raises \texttt{val\_loss} only from 3.47 to 3.57 (small) while crashing \texttt{probe\_in} from 0.98 to 0.40. The heads are doing predominantly probe-task work, not general LM.

\item \textbf{Spectral pre-identification works.} The four heads were identified purely by tracking PR at the query position -- no behavioral labels, no ablation needed. The ablation experiment is a post hoc validation, not the discovery method.
\end{enumerate}

\subsection{Cross-seed: s42 vs s271 asymmetry}
\label{sec:s42_vs_s271}

The full TS-51M panel is six seeds, each with a different spectral signature for the same task:

\begin{center}
\small
\begin{tabular}{lll}
\toprule
Seed & Spectral picks (PR-spread top set) & Where \\
\midrule
s42  & L0H\{3, 6, 14, 15\} & L0 only; every other head has spread $< 12$ \\
s271 & L6H\{1, 10\} + L7H\{9, 15\} & Late layers; no L0 head exceeds spread 11 \\
s149 & L6H\{2, 5, 6, 7\} + L7H\{13\} & Late layers, different specific heads than s271 \\
s256 & L5H10 + L6H\{2, 4\} + L7H\{6, 13\} & Spans L5/L6/L7 (shares L6H2 + L7H13 with s149) \\
s123 & L5H5 + L6H\{5, 11\} + L7H\{2, 4, 13\} & Spans L5/L6/L7 (shares heads with s149, s256) \\
s314 & L5H\{7, 14, 15\} + L7H\{0, 5\} & L5+L7 only (no L6 picks), distinct from all others \\
\bottomrule
\end{tabular}
\end{center}

The spectral signal points at a \emph{different} small set of heads on each seed; PR-spread values for the picks are 20--24 on s42 and 9--11 on s271/s149, while every non-pick head has spread $\leq 14$. The signal-to-noise gap is wide.

\paragraph{Cross-ablation reveals a clean asymmetry.} We run the s42 ablation pipeline with the s271 candidate heads, and vice versa.

\paragraph{s271 ablation (baseline $\mathrm{pin}{=}0.526$ at step~2000):}
\begin{center}
\small
\begin{tabular}{lrr}
\toprule
Condition & pin & $\Delta$ \\
\midrule
baseline                                                & 0.526 & --- \\
ablate s271 circuit (L6H\{1,10\} + L7H\{9,15\})         & 0.273 & $-0.25$ \\
\textbf{ablate s42 circuit on s271 (L0H\{3,6,14,15\})}  & \textbf{0.251} & \textbf{$-0.28$} \\
ablate matched random L6+L7                             & 0.518 & $-0.01$ \\
\bottomrule
\end{tabular}
\end{center}

\paragraph{s42 ablation (multiple checkpoints):}
\begin{center}
\footnotesize
\begin{tabular}{lrrr}
\toprule
Condition & step~800 pin & step~4000 pin & step~10000 pin \\
\midrule
baseline                                                       & 0.982 & 0.843 & 0.995 \\
\textbf{ablate s42 circuit (L0H\{3,6,14,15\})}                 & \textbf{0.396} & \textbf{0.151} & \textbf{0.173} \\
\textbf{ablate s271 circuit on s42 (L6H\{1,10\}+L7H\{9,15\})}  & \textbf{0.982} & \textbf{0.838} & \textbf{0.995} \\
ablate matched random L6+L7                                    & 0.982 & 0.844 & 0.995 \\
ablate ALL 32 heads in L6+L7                                   & 0.982 & 0.826 & 0.941 \\
\bottomrule
\end{tabular}
\end{center}

\textbf{The asymmetry:}

\begin{itemize}[leftmargin=*,topsep=2pt,itemsep=2pt]
\item \textbf{s42's circuit is genuinely L0-localized.} No subset of L6+L7 heads carries probe-task work. Even removing all 32 heads in L6+L7 leaves \texttt{probe\_in\_acc} at 0.94. L6 and L7 are doing other (general LM) computation on this seed.
\item \textbf{s271's circuit is distributed across L0 + L6/L7.} Both subsets drop \texttt{probe\_in\_acc} by similar amounts. The L0H\{3,6,14,15\} heads from s42 are \emph{also} causally relevant on s271.
\item \textbf{Both seeds use L0H\{3,6,14,15\}} -- these are common circuit heads. Only s271 additionally recruits L6/L7 heads to do the same retrieval work.
\end{itemize}

s271's broader circuit correlates with better out-of-distribution generalization (\texttt{probe\_ood} reaches 0.66 vs s42's 0.33). The same architecture trained from different seeds finds different mechanistic implementations of the same task; the shallower L0-only circuit on s42 generalizes worse OOD than the broader L0~+~L6/L7 circuit on s271.

\subsection{Mechanistic characterization: circuit heads attend to the KEY position}
\label{sec:s42_mechinterp}

To characterize \emph{what} the spectrally-identified heads compute, we measure where they attend at the query-read position (the token whose prediction is the first codeword token of the QUERY answer). For each L0 head and each of 200 probe examples, we record the head's softmaxed attention from the query-read position back across the prefix, then sum the attention falling on the position(s) of the KEY codeword's first occurrence.

L0 per-head attention to the KEY position, averaged over 200 probe examples (s42, checkpoint 4000):

\begin{center}
\small
\begin{tabular}{lrrr}
\toprule
Head & attn $\to$ KEY & attn $\to$ self & attn $\to$ uniform-other (baseline) \\
\midrule
L0H0   & 0.041 & 0.032 & 0.004 (control) \\
L0H1   & 0.050 & 0.031 & 0.004 (control) \\
L0H2   & 0.008 & 0.109 & 0.004 \\
\textbf{L0H3}  & \textbf{0.146} & 0.017 & 0.003 \ \  $\leftarrow$ circuit \\
L0H4   & 0.038 & 0.016 & 0.004 \\
L0H5   & 0.041 & 0.042 & 0.004 (control) \\
\textbf{L0H6}  & \textbf{0.139} & 0.027 & 0.003 \ \  $\leftarrow$ circuit \\
L0H7   & 0.027 & 0.068 & 0.004 (control) \\
L0H8   & 0.041 & 0.029 & 0.004 \\
L0H9   & 0.049 & 0.083 & 0.003 \\
L0H10  & 0.011 & 0.059 & 0.004 \\
L0H11  & 0.041 & 0.018 & 0.004 \\
L0H12  & 0.062 & 0.025 & 0.004 \\
L0H13  & 0.049 & 0.046 & 0.004 \\
\textbf{L0H14} & \textbf{0.228} & 0.011 & 0.003 \ \  $\leftarrow$ circuit \\
\textbf{L0H15} & \textbf{0.268} & 0.022 & 0.003 \ \  $\leftarrow$ circuit \\
\bottomrule
\end{tabular}
\end{center}

\paragraph{Selectivity ratio} = $\mathrm{attn}(\to\!\text{KEY}) \,/\, \mathrm{attn}(\to\!\text{uniform other position})$:

\begin{itemize}[leftmargin=*,topsep=2pt,itemsep=2pt]
\item Circuit heads L0H\{3, 6, 14, 15\}: \textbf{42$\times$, 43$\times$, 76$\times$, 95$\times$} more selective for KEY than uniform.
\item All other L0 heads: 2$\times$ -- 18$\times$ (most below 17).
\item Random control heads (L0H\{0, 1, 5, 7\}): 7$\times$ -- 14$\times$.
\end{itemize}

The four spectrally-identified circuit heads are \emph{attending back to the position where the codeword was first mentioned}. Their behavior matches the pattern of induction-style retrieval heads \citep{olsson2022induction}: at a query position they look back through the context for relevant prior content, and the retrieved content is then written into the residual stream via the head's output projection.

\textbf{This explains the spectral signature.} Pre-emergence (step~400) the heads attend to a single default position (likely BOS or the most recent token) regardless of probe content; the V output is concentrated and PR $\approx 2$. Post-emergence (step~800+) the attention has become content-dependent (for each example the head attends to wherever the relevant codeword appeared); since 512 different codewords appear at different positions with different identities, the per-example V output spans many directions and PR $\approx 22$. The PR transition is the spectral signature of the QK circuit becoming content-dependent.

\subsection{What the TS-51M deep dive contributes}

This six-seed validation establishes three things that the larger-scale results in \S\ref{sec:karpathy}--\S\ref{sec:cross_arch} build on:

\begin{enumerate}[leftmargin=*,topsep=2pt,itemsep=2pt]
\item \textbf{The complete spectral $\to$ causal $\to$ mechanistic chain is empirically verified at small scale.} Step~1 (spectral) identifies the right heads, Step~3 (causal) confirms them, and a direct attention-pattern measurement at the same heads gives them a mechanistic story (KEY-position retrieval). The chain holds without label supervision at any step.

\item \textbf{The recipe finds seed-specific circuits, not a universal circuit.} Across six pretraining seeds with identical architecture, hyperparameters, and task, the recipe identifies a different small head set on each seed -- including a striking L0-versus-L6/L7 layer asymmetry between s42 and s271. The recipe is finding the model's actual implementation, not a fixed pattern.

\item \textbf{The matched-random control is informative even when the screen-picked set is tiny.} Across 12 ablation conditions (six seeds $\times$ at least two conditions per seed), the matched-random control consistently lands near zero effect on probe accuracy while the screen-picked set lands near full destruction. The control is what makes the small-set causal claim falsifiable.
\end{enumerate}

\section{Cross-Scale Validation on Natural Text}
\label{sec:cross_scale}

The TS-51M results show the recipe works at small scale on a synthetic probe. The cross-scale validation in this section shows it works on natural-text-pretrained models at 124M, 160M, and 410M parameters, with the same procedure and a consistent 17--19\% conserved fraction.

\subsection{Karpathy 124M deep dive}
\label{sec:karpathy}

\paragraph{Setup.} Karpathy 124M is a 12L $\times$ 768d $\times$ 12h GPT-2-style model trained on FineWeb-10B for 17,600 steps with 89 saved checkpoints \citep{karpathy2023nanogpt}. The synthetic induction batch from \S\ref{sec:setup} is the eval set.

\paragraph{Top-30 picks by PR-integral.} We rank all 144 (layer, head) pairs by Eq.~\ref{eq:integral} and classify the top 30 into the six standard classes at the $\geq 30\times$ threshold:

\begin{center}
\small
\begin{tabular}{rr}
\toprule
$k$ & precision (classified / $k$) \\
\midrule
1, 5, 10, \textbf{15} & \textbf{100\%} \\
20 & 95\% \\
25 & 96\% \\
30 & \textbf{93\%} (28 of 30 classified) \\
\bottomrule
\end{tabular}
\quad
\begin{tabular}{lr}
\toprule
Class & Count (top-30) \\
\midrule
self (attn to current position) & 14 \\
previous-token & 9 \\
induction & 5 \\
unclassified & 2 \\
\bottomrule
\end{tabular}
\end{center}

The two unclassified heads (L1H10, L9H8) have max selectivity 11$\times$--13$\times$ -- weakly content-dependent diffuse heads with no specific dominant pattern.

\paragraph{Causal verification.} Ablate top-6 spectral picks on the final checkpoint and measure induction prediction accuracy on the 2000-example eval:

\begin{center}
\small
\begin{tabular}{lr}
\toprule
Condition & top-1 acc \\
\midrule
baseline & \textbf{16.1\%} \\
\textbf{ablate top-6 spectral picks} (L8H\{8,10,5\}, L6H10, L1H\{9,11\}) & \textbf{0.85\%} \\
ablate matched-random control & 10.6\% \\
upper bound (full spectral-pick layers) & 0\% \\
\bottomrule
\end{tabular}
\end{center}

Spectral-pick ablation drops top-1 by 15.3pp, about 4$\times$ larger than the matched-random control. Individual ablations of the 3 mech-interp-confirmed induction heads (L8H\{8,10,5\}) each drop top-1 by 5--10pp; the 3 false-positive picks (L6H10, L1H\{9,11\}, which are actually prev-token heads, not induction) each drop top-1 by $\leq 1$pp. The spectral signal's top-$K$ can contain capability-class false-positives whose causal role is upstream of the target capability, but the causal effect is dominated by the true-positives.

\paragraph{Robustness: classifications hold across query positions.} A reasonable concern is that classification was done at one query position (the last one). We re-run the classification at five query positions $\{50, 100, 150, 200, 255\}$:

\begin{center}
\small
\begin{tabular}{lrrr}
\toprule
Class @ $p=255$ & Heads & Consistent across 5 positions & Rate \\
\midrule
self & 14 & 11 & \textbf{79\%} \\
previous-token & 9 & 7 & \textbf{78\%} \\
induction & 5 & --- (intrinsically position-specific) & --- \\
\bottomrule
\end{tabular}
\end{center}

The self class is real (79\% consistency, comparable to prev-token's 78\%). About 20\% of single-position labels are position-specific noise (mostly self-vs-prev-token confusion), fixable by multi-position classification.

\paragraph{Side observation: time-of-emergence by class.} For each top-30 pick, when does it first cross PR$=15$?

\begin{center}
\small
\begin{tabular}{lrrl}
\toprule
Class & $n$ & mean step & range \\
\midrule
\textbf{induction} & 5 & \textbf{840} & 800--1000 (very tight) \\
previous-token & 9 & 1556 & 800--4600 (wide) \\
self & 14 & 1257 & 800--2400 \\
\bottomrule
\end{tabular}
\end{center}

Induction heads emerge in a narrow $\sim$200-step window, consistent with the phase-transition character that \citet{olsson2022induction} observed. The companion paper~\citep{paper2_developmental} develops this temporal structure further.

\subsection{Robustness to mid-training data-distribution shift (FineWeb $\to$ OWT)}
\label{sec:fineweb_to_owt}

Karpathy~124M was pretrained on FineWeb-10B through step 17600 (89 checkpoints), then continued on OpenWebText (OWT) starting step 17800. This rare single-model, single-architecture, single-optimizer mid-training data shift isolates the data-distribution component of the recipe's identification claims. We ran the per-head spectral pipeline plus the all-head capability-specific screen at every available checkpoint in both phases.

\begin{figure}[H]
  \centering
  \includegraphics[width=\linewidth]{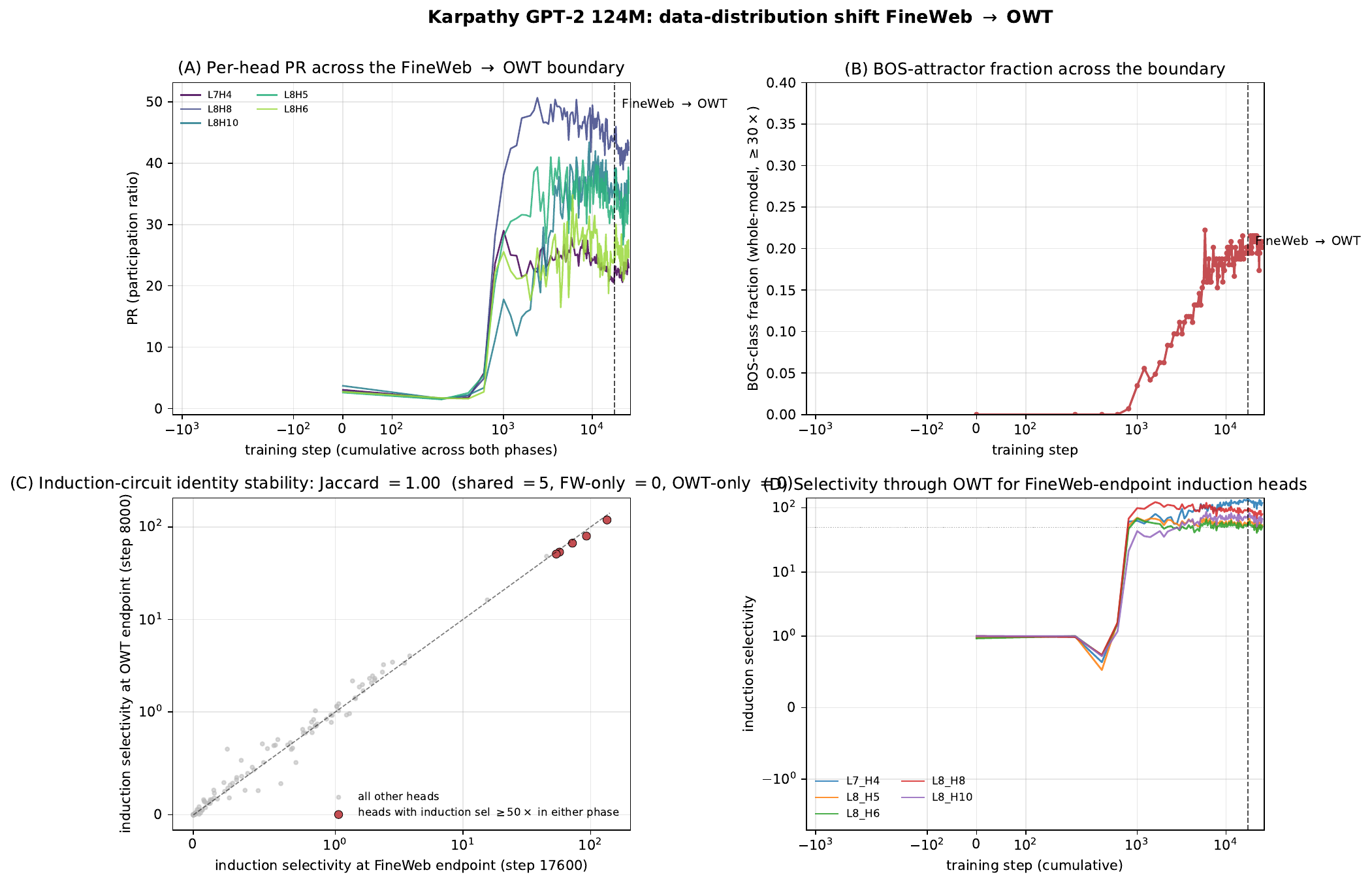}
  \caption{\textbf{The induction circuit identified on FineWeb is invariant to OWT continuation training.} \textbf{(A)}~Per-head PR trajectory across both phases (FineWeb steps 0--17600, OWT steps 0--8000). The five FineWeb-endpoint induction-circuit heads (L8$\cdot$H8, L8$\cdot$H5, L8$\cdot$H6, L8$\cdot$H10, L7$\cdot$H4) maintain elevated PR across the boundary; no head transitions out of the circuit. \textbf{(B)}~Whole-model BOS-class fraction (fraction of heads classified first-token at $\geq 30\times$) across the boundary. The fraction is 0.201 at FineWeb step 17600 and 0.201 at OWT step 8000 -- BOS-attractor formation is independent of the FineWeb-vs-OWT data distribution. \textbf{(C)}~Induction-circuit identity stability: scatter of induction-selectivity at FineWeb endpoint (x-axis) vs OWT endpoint (y-axis). Heads at $\geq 50\times$ in either phase are highlighted; \textbf{Jaccard = 1.00} (shared 5, FineWeb-only 0, OWT-only 0). \textbf{(D)}~Per-head capability selectivity through OWT for the five FineWeb-endpoint induction heads. Selectivity is approximately constant; no decay, no replacement.}
  \label{fig:fineweb_to_owt}
\end{figure}

Four measurements, all consistent with circuit invariance across the data shift:

\begin{itemize}[leftmargin=*,topsep=2pt,itemsep=2pt]
\item \textbf{PR continuity (Figure~\ref{fig:fineweb_to_owt}A):} the per-head PR trajectory is continuous across the FineWeb $\to$ OWT boundary; the five FineWeb-endpoint induction heads maintain elevated PR throughout the OWT continuation.
\item \textbf{Induction-circuit Jaccard $=1.00$ (Figure~\ref{fig:fineweb_to_owt}C):} the same five heads (L8$\cdot$H\{5, 6, 8, 10\}, L7$\cdot$H4) are at $\geq 50\times$ induction-selectivity at both the FineWeb endpoint (step 17600) and the OWT endpoint (step 8000). No identity migration.
\item \textbf{BOS-fraction flat (Figure~\ref{fig:fineweb_to_owt}B):} whole-model BOS-class fraction at $\geq 30\times$ is 0.201 at FineWeb step 17600 and 0.201 at OWT step 8000 -- identical to three decimal places.
\item \textbf{Per-head selectivity stable (Figure~\ref{fig:fineweb_to_owt}D):} the FineWeb-identified circuit's induction-selectivity is approximately constant through the full OWT continuation; no decay, no replacement.
\end{itemize}

\paragraph{Reading.} The induction circuit identified by the recipe on a model trained on one corpus does not migrate to different heads, does not lose selectivity, and does not face attention-sink reorganization when the model is continued on a similar-but-different corpus. The methodology is identifying \emph{structural specialization} rather than \emph{corpus statistics} -- a strong robustness claim that the synthetic-batch evaluation alone cannot make. Caveats: FineWeb and OWT are both English web text, the OWT continuation is short ($\sim$0.5B tokens), and Karpathy~124M was already well-trained when the data switched. A more distant shift (code, multilingual, scientific) or a longer continuation might surface drift this test cannot.

\subsection{Integral vs spread: nine ranking signals compared}
\label{sec:integral_vs_spread}

We tested nine alternative trajectory features against the PR-integral for ranking. The PR-integral wins on Karpathy 124M:

\begin{center}
\small
\begin{tabular}{lrr}
\toprule
Ranking signal & Precision-at-30 (Karpathy 124M) & Mean selectivity in top-5 \\
\midrule
spread (max $-$ min) & 0.93 & 155$\times$ \\
\textbf{integral} & \textbf{0.97} & \textbf{5{,}791}$\times$ \\
max\_pr & 0.93 & 177$\times$ \\
mean\_post\_grok & 0.93 & 258$\times$ \\
max\_rate & 0.53 & 169$\times$ \\
\bottomrule
\end{tabular}
\end{center}

(The full nine-feature table includes additional variants that all underperform integral; we report the five most representative.)

Why integral wins: it rewards \emph{sustained} high PR, whereas spread only measures the max-minus-min gap. On Karpathy 124M, L6H9 (27{,}776$\times$ previous-token selectivity) is rank~14 by spread but rank~5 by integral -- spread underrates heads that are consistently elevated.

\textbf{Integral is essential on Pythia.} L0 heads on Pythia start at PR $\approx 60$ (random attention at init produces a high effective rank) and \emph{collapse} to PR $\approx 2$--$30$ by training end. PR-spread flags them as top picks; PR-integral correctly demotes them in favor of heads that gain sustained PR through training.

\subsection{Pythia 160M and 410M: precision-at-k across 8x scale}
\label{sec:pythia_p_at_k}

We run the same pipeline on EleutherAI Pythia 160M (12L $\times$ 768d $\times$ 12h, dense, Pile) and Pythia 410M (24L $\times$ 1024d $\times$ 16h, dense, Pile) to test whether the spectral signal generalizes across (data, training procedure, RNG, codebase, scale).

\begin{center}
\small
\begin{tabular}{rrrr}
\toprule
$k$ & Karpathy 124M (FineWeb) & Pythia 160M (Pile) & Pythia 410M (Pile) \\
\midrule
5  & 100\% & 100\% & 100\% \\
10 & 100\% & 100\% & 100\% \\
15 & 100\% & 93\%  & 93\% \\
30 & 93\%  & 93\%  & 90\% \\
50 & ---   & ---   & 90\% \\
80 & ---   & ---   & 81\% \\
\bottomrule
\end{tabular}
\end{center}

Precision-at-$k$ matches within 1--3pp across an 8$\times$ parameter scale range and two completely different training pipelines.

\subsection{The conserved 17--19\% fraction}
\label{sec:conserved_fraction}

Extending Pythia 410M's classification to top-80 (head-count-matched to top-30 on the 144-head models) and counting heads classified into a known capability class:

\begin{center}
\small
\begin{tabular}{lrrrr}
\toprule
Model & Total heads & $k$ (matched) & Classified & Fraction \\
\midrule
Karpathy 124M & 144 & 30 & 28 & \textbf{19.4\%} \\
Pythia 160M   & 144 & 30 & 27 & \textbf{18.8\%} \\
Pythia 410M   & 384 & 80 & 65 & \textbf{16.9\%} \\
\bottomrule
\end{tabular}
\end{center}

\textbf{Roughly 17--19\% of heads in a model do identifiable specialized computation, conserved across an 8$\times$ scale range.} The capability \emph{count} scales with model size; the \emph{fraction} of heads doing specialized work stays constant.

The PR-integral distribution has a natural elbow that scales similarly:

\begin{center}
\small
\begin{tabular}{lrrr}
\toprule
Model & Total heads & Elbow $k$ & Elbow / total \\
\midrule
Karpathy 124M & 144 & 30 & 20.8\% \\
Pythia 160M   & 144 & 23 & 16.0\% \\
Pythia 410M   & 384 & 70 & 18.2\% \\
\bottomrule
\end{tabular}
\end{center}

The elbow sits at $\sim$16--21\% of total heads in every model -- a model-agnostic cutoff that does not require choosing $k$ by hand.

\subsection{Each model has one super-prev-token head}
\label{sec:super_prev}

A surprising regularity is that every model in the panel has exactly one head with extreme prev-token selectivity:

\begin{center}
\small
\begin{tabular}{lll r}
\toprule
Model & Head & prev-token selectivity \\
\midrule
Karpathy 124M & L6H9 & 27{,}776$\times$ \\
Pythia 160M   & L3H2 & \textbf{81{,}792}$\times$ \\
Pythia 410M   & L5H2 & 23{,}634$\times$ \\
\bottomrule
\end{tabular}
\end{center}

Different layer in each model, but the integral ranking finds it across all three (PR-spread misses all three -- their trajectories are high-but-not-the-highest-spread). This suggests a real architectural regularity: there is probably one degree of freedom per model that gets compressed into a single near-perfect prev-token implementation, and the methodology is sensitive enough to find it without knowing where to look.

\subsection{Class-mix shifts with scale}

\begin{center}
\small
\begin{tabular}{lrrrr}
\toprule
Class & Karpathy 124M & Pythia 160M & Pythia 410M (top-30) & Pythia 410M (top-80) \\
\midrule
previous-token & 9  & 9 & 14 & 25 \\
self           & 14 & 11 & 9 & 18 \\
induction      & 5  & 2 & 1 & 1 \\
first-token (BOS) & 0 & 6 & 3 & 20 \\
unclassified   & 2  & 2 & 3 & 15 \\
\bottomrule
\end{tabular}
\end{center}

Pythia (Pile) has many more first-token-attending heads than Karpathy (FineWeb), possibly tied to BOS-token usage in Pile. The induction class in top-30 shrinks with scale, but as the next section shows that does not mean induction is \emph{lost} -- it is just distributed across more heads at lower per-head selectivity.

\subsection{Distribution vs dilution: the 11-head all-head induction screen on 410M}
\label{sec:distribution_vs_dilution}

The Pythia 410M induction ablation evolved across three targeting attempts: (a)~top-6 spectral $\to$ 43\% top-1 drop, (b)~mech-interp + 2nd-class $\to$ 77\%, (c)~all-head induction screen $\to$ 100\%. Two hypotheses for why the smaller attempts didn't fully tank induction:

\begin{itemize}[leftmargin=*,topsep=2pt,itemsep=2pt]
\item \textbf{Dilution}: induction at scale has lower per-head selectivity; we genuinely captured less of the circuit.
\item \textbf{Distribution}: induction is spread across more heads at lower individual selectivity; our targeting was incomplete.
\end{itemize}

Screening all 384 heads of Pythia 410M for induction selectivity $\geq 50\times$ found \textbf{11 such heads} -- \emph{more} than Karpathy 124M's 6 -- but at lower per-head selectivity (max 203$\times$ vs Karpathy's 681$\times$).

\begin{center}
\small
\begin{tabular}{lrr}
\toprule
Condition & top-1 & $n$ heads ablated \\
\midrule
baseline & 3.7\% & --- \\
previous ``extended'' set (3 heads) & 0.85\% & 3 \\
\textbf{all heads with induction sel $\geq 100\times$ (8 heads)} & \textbf{0.10\%} & \textbf{8} \\
\textbf{all heads with induction sel $\geq 50\times$ (11 heads)} & \textbf{0.0\%} & \textbf{11} \\
\bottomrule
\end{tabular}
\end{center}

\textbf{Distribution wins.} Total induction ``signal'' is preserved, just spread across more heads. The methodology fully captures the circuit when ablation targets the full induction-selective set. Only 3 of the 11 induction-selective heads on 410M were in the top-30 spectral picks; the other 8 (L8H6 at 203$\times$, L7H1 at 177$\times$, etc.) live at lower integral ranks and would be missed by spectral-picks-only mech-interp. This is the operative justification for the all-head capability-specific screen in \S\ref{sec:step2}.

\paragraph{Multi-purpose heads.} L11H14 on Pythia 410M is primary-classified as first-token (287$\times$) but has strong second-class induction (124$\times$); single-head ablation tanks induction by 50\% on its own. Heads with multiple non-trivial selectivities are common in larger models; second-class selectivities can be load-bearing for capability-specific ablation. The capability-specific all-head screen naturally picks these up.

\paragraph{Causal verification at each scale.} Summarizing the cross-scale ablation results:

\begin{center}
\small
\begin{tabular}{llr}
\toprule
Model & Targeting & top-1 drop \\
\midrule
Karpathy 124M & top-6 spectral picks (incl 3 induction heads) & 16\% $\to$ 0.85\% ($-95$\%) \\
Pythia 160M   & mech-interp-classified L8H2 + L5H0           & 4.7\% $\to$ 0.05\% ($-99$\%) \\
Pythia 410M   & top-6 spectral picks (mostly self/prev)      & 3.7\% $\to$ 2.1\% ($-43$\%) \\
Pythia 410M   & mech-interp-classified + 2nd-class induction & 3.7\% $\to$ 0.85\% ($-77$\%) \\
Pythia 410M   & \textbf{all heads with induction sel $\geq 50\times$ (11 heads)} & \textbf{3.7\% $\to$ 0.0\% ($-100$\%)} \\
\bottomrule
\end{tabular}
\end{center}

\section{Cross-Architecture Induction at 1B-class Scale}
\label{sec:cross_arch}

The cross-scale results in \S\ref{sec:cross_scale} hold across an 8$\times$ parameter range within a single architecture family (dense GPT-style transformers). In this section we extend the recipe across architecture families: Pythia~1B (GPT-NeoX dense, Pile), OLMo 1B-0724-hf (Llama-style dense, DCLM-aligned), and OLMoE 1B-7B-0924 (Llama-style MoE, DCLM). All three are 1B-active-parameter models trained on different data with different codebases.

\subsection{Attention-sink-dominated regime}

These three models are all in the attention-sink-dominated regime: BOS-class fractions at the final checkpoint range from 54\% (Pythia 1B) to 78\% (OLMo 1B). Best-class ranking surfaces BOS-class heads at the top of the spectral integral list. The all-head capability-specific screen (\S\ref{sec:step2}) is the robust approach: screen all heads with induction selectivity $\geq 50\times$ at the final checkpoint, then ablate as a group.

\subsection{Induction circuit ablation across three 1B-class models}

\begin{center}
\footnotesize
\begin{tabular}{llrrr}
\toprule
Model & Induction circuit ($\geq 50\times$) & Baseline & After ablation & Matched-random \\
\midrule
Pythia 1B   & 3 heads: L4H4, L7H0, L7H1 & 4.05\% & 0.25\% ($\Delta -3.80$) & 4.40\% ($\Delta +0.35$) \\
OLMo 1B     & 3 heads: L2H11, L4H12, L12H8 & 1.00\% & 0.05\% ($\Delta -0.95$) & 2.20\% ($\Delta +1.20$) \\
OLMoE 1B-7B & 4 heads: L5H10, L7H0, L9H8, L12H14 & 4.80\% & 0.00\% ($\Delta -4.80$) & 1.30\% ($\Delta -3.50$) \\
\bottomrule
\end{tabular}
\end{center}

The induction circuit is small (3--4 heads, sublinear in model size) and causally necessary for synthetic induction in all three 1B-class models. Matched-random in the same layers has zero or positive effect on top-1 in Pythia 1B and OLMo 1B; the induction-pattern heads carry the signal, not their layer neighbors. OLMoE 1B-7B has a smaller specificity differential (4.80 vs 3.50; 1.4$\times$) because the matched-random control samples from layers (L5, L7, L9, L12) that are densely populated with other induction-adjacent heads.

\subsection{Per-model ablation curves and circuit granularity}
\label{sec:per_model_ablation_curves}

The per-model ablation-floor sweep (\S\ref{sec:threshold_calibration}) at $T \in \{2, 10, 30, 50, 100\}\times$ refines the per-model circuit size:

\begin{itemize}[leftmargin=*,topsep=2pt,itemsep=2pt]
\item \textbf{Pythia 1B (baseline 4.05\%):} $\geq 50\times$ (3 heads) $\to$ 0.25\% (94\% drop); $\geq 30\times$ (6 heads) $\to$ 0.05\% (99\% drop); $\geq 10\times$ (11 heads) $\to$ 0\% (full closure). The full causal circuit extends to $\sim$6 heads; the additional 3 heads in the 30--50$\times$ band are multi-role first-token + induction.
\item \textbf{OLMo 1B (baseline 1.00\%):} $\geq 100\times$ (just \textbf{2 heads}: L2H11 and L4H12) already reaches the ablation floor (0.05\%). Adding heads down to $\geq 30\times$, $\geq 10\times$, $\geq 2\times$ produces no further drop. OLMo~1B's full induction capability is carried by 2 extremely sharp specialists, with matched-random controls showing no other 2-head set in those layers reproduces the effect.
\item \textbf{OLMoE 1B-7B (baseline 4.80\%):} $\geq 30\times$ and $\geq 50\times$ return the \textbf{identical 4 heads} (L5H10, L7H0, L9H8, L12H14). Ablation at this set reaches the floor (0\% top-1). Any threshold in the 30--100$\times$ range identifies the same circuit -- the strongest signature of a well-separated circuit in the panel.
\end{itemize}

\begin{figure}[H]
  \centering
  \includegraphics[width=0.92\linewidth]{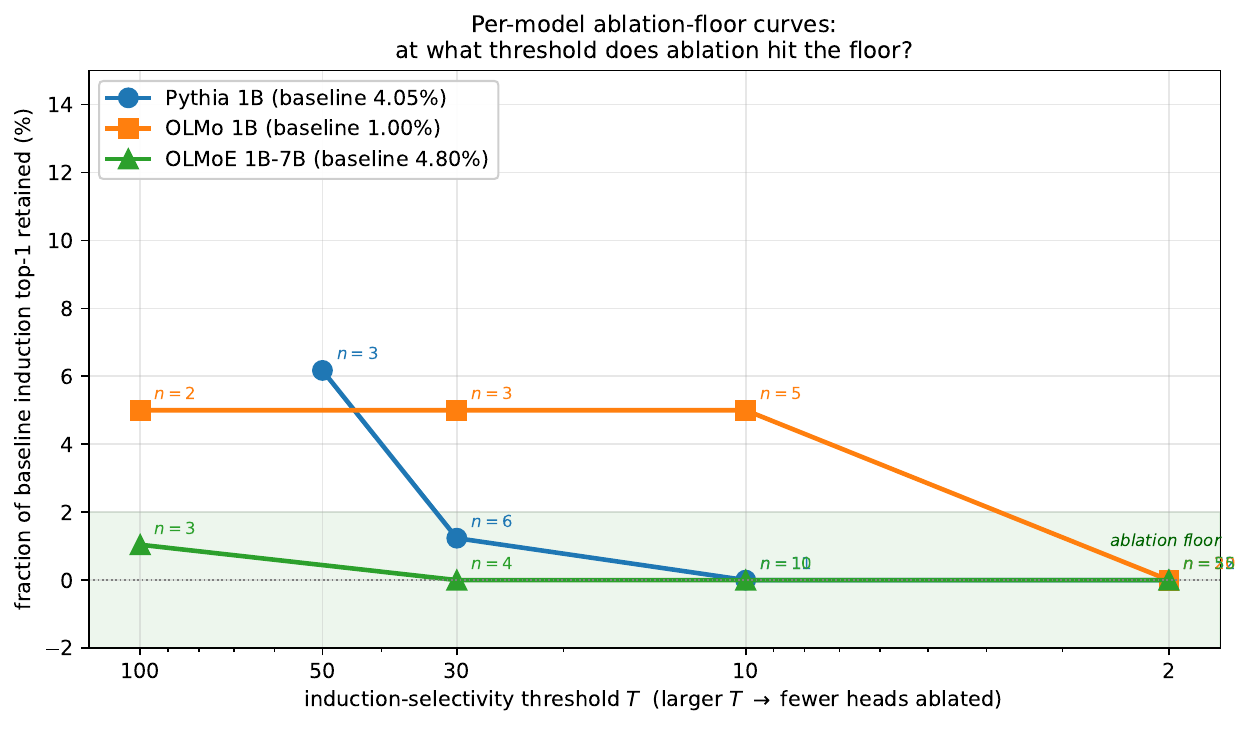}
  \caption{\textbf{Per-model ablation-floor curves: at what threshold $T$ does ablating all heads with induction-selectivity $\geq T\times$ hit the ablation floor?} Y-axis is fraction of baseline synthetic-induction top-1 retained after ablation; x-axis is $T$ (decreasing leftward, so progressively more heads get included as one moves left). \textbf{OLMo 1B (orange):} the curve is already at $\sim 5\%$ retention by $T{=}100$ ($n{=}2$ heads ablated) and stays flat until $T{=}2$, where the floor is reached -- the full causal effect is captured by just two heads at $\geq 100\times$. \textbf{OLMoE 1B-7B (green):} retention is $\sim 1\%$ at $T{=}100$ ($n{=}3$) and reaches the floor at $T{=}30$ ($n{=}4$), then stays flat -- a sharp plateau in the 30--100$\times$ range. \textbf{Pythia 1B (blue):} still retains $\sim 6\%$ at $T{=}50$ ($n{=}3$), drops to $\sim 1\%$ at $T{=}30$ ($n{=}6$), reaches the floor only at $T{=}10$ ($n{=}11$) -- a distributed circuit with substantial contribution from heads in the 10--50$\times$ band. The same task is implemented at three different mechanistic granularities across the panel.}
  \label{fig:ablation_floor_curves}
\end{figure}

\paragraph{Induction-circuit granularity differs across architectures.} OLMo~1B implements induction with 2 extremely sharp specialists ($\geq 100\times$); OLMoE~1B-7B uses 4 mid-selectivity heads on a wide plateau (30--200$\times$ range); Pythia~1B distributes across 6+ multi-role heads in the 10--50$\times$ band. Same task, three different mechanistic granularities of implementation. The three ablation-floor curves are shown together in Figure~\ref{fig:ablation_floor_curves}: OLMo's flat-from-$T{=}100$ line, OLMoE's plateau-from-$T{=}30$ line, and Pythia 1B's still-descending-at-$T{=}50$ line are visibly three different curve shapes.

 This complements the cross-architecture composed-task finding~\citep{paper3_circuits} that the same task is implemented by different attention patterns across models: across the 1B panel, induction differs both in \emph{which patterns} its heads use \emph{and in how many heads} participate at \emph{what selectivity level}.

\subsection{Natural-text confirmation}
\label{sec:naturaltext_induction}

The synthetic-batch-identified circuits also produce a causal effect on natural-text induction. For OLMoE, the 4-head induction circuit ablation on OpenWebText prompts at induction-target positions produces a 7.5$\times$ top-1 differential and a 13.5$\times$ logit-of-target differential over matched-random in the same layers. Same direction on Pythia 410M (synthetic-identified induction heads degrade natural-text loss at induction-target positions). The circuit identified on the synthetic batch is the circuit doing induction on real text.

\subsection{Scope note on the developmental and composed-task extensions}

The recipe also applies at intermediate training checkpoints, where the spectral signal precedes capability-selectivity emergence and the screen recovers most of the final-checkpoint circuit using 0.3--2\% of training tokens. The developmental analysis is the subject of the companion paper~\citep{paper2_developmental}; here we restrict attention to final-checkpoint identification, which is sufficient for the recipe and the cross-architecture replication.

Composed tasks (Indirect Object Identification, greater-than, successor sequences, variable binding) introduce a separate phenomenon: the same task is implemented by different primary attention patterns across model families, and the screen for a composed task does not port across families. That decoupling is developed in detail in the second companion paper~\citep{paper3_circuits}.

\section{Previous-Token Circuit and Compositional Confirmation}
\label{sec:prevtoken}

The all-head capability-specific screen extends straightforwardly to other classes. For previous-token, we filter heads where best-class is previous-token AND selectivity $\geq 100\times$ (a tighter threshold than the induction circuit because prev-token selectivity values are larger in absolute terms -- some heads reach $10^6$--$10^7$ selectivity).

\subsection{Prev-token circuit identification}

\begin{center}
\footnotesize
\begin{tabular}{lll}
\toprule
Model & Prev-token circuit & Top heads (head, prev-sel) \\
\midrule
Pythia 1B   & 13 ($\geq 50\times$), 8 ($\geq 100\times$) & L3H5 ($9.8\!\times\!10^7$), L6H3 ($1.1\!\times\!10^6$), L3H6 (770), L1H3 (374) \\
OLMo 1B     & 11 ($\geq 50\times$), 9 ($\geq 100\times$) & L11H10 (10477), L0H5 (654), L1H2 (486), L0H6 (239) \\
OLMoE 1B-7B & 12 ($\geq 50\times$), 8 ($\geq 100\times$) & L8H10 (5178), L0H1 (2369), L6H6 (2088), L8H13 (605) \\
\bottomrule
\end{tabular}
\end{center}

The prev-token circuit is 2--4$\times$ the size of the induction circuit (8--9 heads vs 3--4). Prev-token specialists concentrate in early layers (L0--L3 dominant), with the strongest individual specialists in Pythia 1B's L3H5 (prev-token selectivity $\sim 10^8$) and OLMo's L11H10 ($\sim 10^4$). These are heads that essentially do nothing except attend to position $t-1$.

\subsection{Compositional confirmation: prev-token ablation tanks induction}

Ablating the prev-token circuit ($\geq 100\times$ threshold, 8--9 heads) and measuring the effect on \emph{synthetic induction} top-1:

\begin{center}
\footnotesize
\begin{tabular}{lrrrr}
\toprule
Model & Prev-token circuit & Induction baseline & After ablation & Matched-random \\
\midrule
Pythia 1B   & 8 heads & 4.05\% & 0.00\% ($\Delta -4.05$) & 4.40\% ($\Delta +0.35$) \\
OLMo 1B     & 9 heads & 1.00\% & 0.00\% ($\Delta -1.00$) & 2.20\% ($\Delta +1.20$) \\
OLMoE 1B-7B & 8 heads & 4.80\% & 0.30\% ($\Delta -4.50$) & 1.30\% ($\Delta -3.50$) \\
\bottomrule
\end{tabular}
\end{center}

Prev-token-circuit ablation tanks induction across the panel. This is the standard compositional structure \citep{olsson2022induction}: prev-token heads build the K-vectors that induction heads use to point at the position after the earlier occurrence. The all-head screen recovers the prev-token compositional dependency from selectivity alone, without separately identifying induction-versus-prev-token computational role.

\paragraph{Methodological note.} Two screens (induction-selective $\geq 50\times$ and prev-token-best-class $\geq 100\times$) both produce circuits that tank synthetic-induction top-1 to near zero. They are not the same circuit (no head overlap in Pythia 1B; partial overlap in OLMo and OLMoE). Both identifications are valid; the prev-token screen catches the compositional upstream component, the induction screen catches the output component. The recipe's matched-random differential is what guarantees each is a real causal claim and not an artifact of layer choice.

\section{Cross-Panel Invariants}
\label{sec:invariants}

Two cross-panel invariants are documented here, restricted to those that fall within the methodology paper's scope. (Developmental invariants involving training-trajectory structure are deferred to the companion paper~\citep{paper2_developmental}.)

\subsection{Conserved 17--19\% specialized-computation fraction across 8x scale}
\label{sec:invariant_fraction}

Restating from \S\ref{sec:conserved_fraction}: across an 8$\times$ parameter range (124M to 410M; the 1B-class models occupy the BOS-dominated regime that requires the all-head screen rather than top-$k$ classification), the fraction of heads classified into a known capability class at the matched top-$k$ stays in the 17--19\% band:

\begin{center}
\small
\begin{tabular}{lrrrr}
\toprule
Model & Total heads & $k$ (matched) & Classified & Fraction \\
\midrule
Karpathy 124M & 144 & 30 & 28 & 19.4\% \\
Pythia 160M   & 144 & 30 & 27 & 18.8\% \\
Pythia 410M   & 384 & 80 & 65 & 16.9\% \\
\bottomrule
\end{tabular}
\end{center}

Capability head \emph{count} scales linearly with total head count; the \emph{fraction} of model capacity used for specialized work is conserved.

\paragraph{Independent recovery by null-selectivity calibration.} The conserved fraction was originally identified by classifying top-$k$ heads into capability classes at $\geq 30\times$ selectivity. An independent calibration procedure -- counting heads with induction-selectivity above null$_{p99}$, where the null is the same metric computed against a random non-special target position (\S\ref{sec:threshold_calibration}) -- recovers the same band on Pythia~160M (18.1\%) and Pythia~410M (18.5\%) without targeting it. Two methodologically distinct procedures producing the same fraction is strong evidence that the 17--19\% band is a structural property of natural-text Pythia models, not an artifact of either classification rule. The band does \emph{not} generalize cross-architecture (Pythia 1B 25.8\%, OLMo~1B 4.7\%, OLMoE 27.3\% under the same null-selectivity rule); the conservation is within-family, not universal.

\subsection{Sublinear circuit-size scaling}

Induction circuits stay 3--11 heads from 124M to 1B-active-parameter MoE:

\begin{center}
\small
\begin{tabular}{lrr}
\toprule
Model & Induction circuit size & Total heads \\
\midrule
GPT-2 124M (Karpathy) & 3--6 (depending on threshold) & 144 \\
Pythia 410M           & 11 (all-head screen) & 384 \\
Pythia 1B             & 3 & 128 \\
OLMo 1B               & 3 & 256 \\
OLMoE 1B-7B           & 4 & 256 \\
\bottomrule
\end{tabular}
\end{center}

The induction circuit does not scale linearly with the number of attention heads. A small fixed circuit of 3--11 heads carries induction in dense models from 124M to 1B; the MoE model uses 4 heads at 1B-active~/~7B-total. This is consistent with the conserved-fraction observation above (and refines it: induction specifically uses far fewer heads than the conserved-fraction band).

The companion paper~\citep{paper2_developmental} documents two further cross-panel invariants -- the L0/L1 zero-BOS architectural floor and the BOS-fraction scaling with training data and architecture -- that are developmental rather than identification-procedure findings; they are reported there rather than here.

\section{Discussion}

\subsection{What the recipe is}

A workflow for identifying small attention-head circuits in transformers, suitable for use during pretraining (Step~1's spectral signal is read off per-checkpoint) and at fully-trained models alike (Steps~2--3 work on any single checkpoint). The recipe operates per-head, requires no labels or attribution gradients, and produces verifiable circuit identifications via standard ablation.

The recipe is a \emph{recipe}, not a universal screen. Step~1 (spectral signal) is universal across the panel; the same PR-integral computation surfaces specialized heads in every model tested. Step~2 (task-pattern screen) is task-specific: the standard six-class capability screen plus task-specific patterns as needed. Step~3 (causal verification) is a control structure (matched-random in the same layers) rather than a specific test. The combination ports across architectures and training pipelines.

\subsection{What the recipe is not}

The methodology does \emph{not} claim that the spectral signal alone identifies task-specific circuits. PR-integral top-$K$ is a general specialization indicator -- in attention-sink-dominated 1B-class models, the top-$K$ is dominated by L0/L1 generic content-dependent heads, not the induction or prev-token circuits. The task-pattern screen does the task-specific work; the spectral signal can be thought of as a fast pre-filter for ``where is interesting computation happening at all.''

The methodology does \emph{not} claim that a task-pattern screen identifying a circuit on one model identifies that circuit on another. For composed tasks (IOI, greater-than, successor, variable binding), the second companion paper~\citep{paper3_circuits} demonstrates that the primary attention pattern carrying causal effect differs across model families -- four different task-relevant attention patterns map to three different primary causal circuits on the same task. A complete pretrained-LM circuit map per task needs a family of candidate screens plus per-model causal validation.

\subsection{Limitations}

\begin{itemize}[leftmargin=*,topsep=2pt,itemsep=2pt]
\item \textbf{Eval batch dependence.} Both the spectral signal and the capability screen are computed on a single fixed evaluation batch (RNG seed 42). Sensitivity to batch composition has been spot-checked (cross-position consistency on Karpathy 124M: 78\% prev-token, 79\% self, \S\ref{sec:karpathy}) but not systematically characterized for the 1B-class models.

\item \textbf{Capability class coverage.} The 6-class standard set covers a small set of attention patterns. Other capability classes (e.g., negative name-movers, copy suppression, backup heads) are not screened for in this work. They appear in the composed-task analysis of the companion paper~\citep{paper3_circuits}.

\item \textbf{Causal target restricted to top-1 / logit\_diff.} Ablation effects are measured on top-1 accuracy and target logit differential. More granular metrics (per-example logit attribution, distribution-shift sensitivity) are not reported. The companion composed-task paper~\citep{paper3_circuits} shows that top-1 vs logit\_diff can disagree (S-Inhibition shifts logit\_diff in OLMoE 1B-7B but not top-1) and that finer-grained metrics are likely needed for circuits whose causal effect is at the margin rather than at the argmax.

\item \textbf{Single seed per model except TS-51M.} The natural-text models (124M, 160M, 410M, 1B-class) are single pretrained checkpoints. Whether the \emph{specific} circuit heads are the same across re-pretrains with different seeds is not tested at scale. The TS-51M six-seed experiment shows that the specific heads differ across seeds even on the same task; the \emph{spectral signal} identifies the correct seed-specific heads on each seed. Whether this also holds for natural-text-pretrained models is an open question; the conserved 17--19\% fraction across 124M~/~160M~/~410M is a circumstantial argument for some kind of type-level stability.

\item \textbf{BOS-saturated regime caveat.} In models where $\geq 70\%$ of heads classify as first-token at the standard threshold (Pythia 410M, OLMo 1B, OLMoE 1B-7B), the PR-integral top-$K$ is dominated by L0/L1 generic content-dependent heads rather than the task-specific circuit. The all-head capability-specific screen (\S\ref{sec:step2}) is what makes the recipe work in this regime; using top-$K$ alone would surface BOS-class heads ahead of induction or prev-token specialists.

\item \textbf{MoE forward-pass cost limits granularity.} Per-revision mech-interp on OLMoE 1B-7B is the slowest in the panel because the model loads 7B parameters even when active inference uses 1B.
\end{itemize}

\subsection{Practical recommendation}

For a researcher wanting to apply the recipe to a new task on a new model:

\begin{enumerate}[leftmargin=*,topsep=2pt,itemsep=2pt]
\item Run Step~1 (PR-integral per head) on the model's training trajectory if available, or at the final checkpoint over multiple eval-batch sub-samples otherwise. This gives a list of heads doing specialized computation.
\item List the attention patterns the task plausibly requires. For composed tasks, this list may have 3--5 entries.
\item Run Step~2 (all-head selectivity screen) for each candidate pattern. For each, identify a small circuit ($\leq 15$ heads, threshold-based).
\item Run Step~3 (group ablation + matched-random + upper bound) on each candidate circuit. Report the differential vs matched-random in the same layers.
\item Accept the screen with the largest specificity differential (target ablation $\Delta$ relative to matched-random $\Delta$) as the primary circuit. The next-largest screen, if its differential is meaningful ($\geq \sim 3\times$), is a real secondary mechanism.
\end{enumerate}

\paragraph{Reading guide for the trilogy.} This paper establishes Steps~1--3 and validates them on single-pattern capabilities (induction, previous-token) across an 8$\times$ scale range and three architecture families. The companion paper on developmental trajectories~\citep{paper2_developmental} uses the same recipe to study circuit formation \emph{during} pretraining and documents that the spectral signal precedes capability-selectivity emergence. The companion paper on composed tasks~\citep{paper3_circuits} uses the same recipe to study IOI, greater-than, successor, and variable binding across the 1B-class panel, and shows that the primary attention pattern carrying causal effect differs across model families.

\appendix

\section{Matched-random seed sensitivity}
\label{app:matched_random_seed}

The matched-random control is reported throughout this paper at a single fixed seed (seed=123) per condition. To check whether this single-seed value is representative, we re-ran the $\geq 10\times$ induction-screen matched-random ablation with 8 additional seeds on each 1B-class model, mirroring the original per-layer head-count distribution. Each ablation is otherwise identical (same induction batch with seed-42 batch generation, same eval procedure).

\begin{center}
\footnotesize
\begin{tabular}{lrrrrrrr}
\toprule
Model & baseline & seed=123 (orig) & 8-seed mean & std & min & max & orig's quantile \\
\midrule
Pythia 1B   & 4.05\% & \textbf{36.05\%} & 8.02\%  & 8.20pp & 0.65\% & 28.30\% & \textbf{1.0} (top) \\
OLMo 1B     & 1.00\% & 0.90\%           & 1.12\%  & 0.40pp & 0.75\% & 1.95\%  & 0.50 (median) \\
OLMoE 1B-7B & 4.80\% & 6.55\%           & 6.73\%  & 3.49pp & 3.85\% & 15.60\% & 0.75 \\
\bottomrule
\end{tabular}
\end{center}

OLMo 1B's matched-random distribution is tightly concentrated around baseline (std 0.40pp, range $\sim$1pp), and the originally reported seed=123 value sits at the median --- single-seed reporting is representative. OLMoE 1B-7B has moderate variance (std 3.49pp, range $\sim$12pp); the originally reported 6.55\% is at the 75th percentile, still within the bulk of the distribution.

Pythia 1B is qualitatively different: the originally reported 36.05\% is at the top of the 9-seed distribution and the 8-seed mean is 8\%. Three additional isolation controls (sink-only ablation in the same layers; isolated ablation of the two non-overlap heads in the seed=123 sample; isolated ablation of the single multi-class head in that sample) failed to reproduce a comparable release on their own, and removing those candidate heads from the seed=123 matched-random set did not eliminate the release. We could not identify a consistent mechanism. The matched-random release in Pythia 1B's induction-circuit layers is dominated by seed variance.

\paragraph{Recommendation.} For matched-random conditions where the effect is small and one-sided (the typical case in this paper: matched-random near baseline, confirming screen specificity), single-seed reporting is adequate. Where the effect is large or has unexpected sign, multi-seed sampling ($\geq 5$ seeds) is necessary; only the Pythia 1B large-effect case in this paper required the sweep.

Raw data is available in the companion repository.

\bibliography{refs}

\end{document}